\documentclass[sn-mathphys-num]{sn-jnl}

\usepackage{graphicx}
\graphicspath{{figures/}}%
\usepackage{color}
\usepackage{enumitem}
\usepackage{multirow}
\usepackage{multicol}
\usepackage{array}
\usepackage{subcaption,ragged2e}
\usepackage{framed, amsmath, amssymb}
\usepackage{booktabs}
\newcolumntype{C}[1]{>{\centering\arraybackslash}p{#1}}

\usepackage{adjustbox}

\newcommand{\new}[1]{#1}

\begin{document}

\title[Autograding Mathematical Induction Proofs with Natural Language 
Processing]{Autograding Mathematical Induction Proofs with Natural Language Processing}


\author*[1]{\fnm{Chenyan} \sur{Zhao}}\email{chenyan4@illinois.edu}

\author[1]{\fnm{Mariana} \sur{Silva}}\email{mfsilva@illinois.edu}

\author[2]{\fnm{Seth} \sur{Poulsen}}\email{seth.poulsen@usu.edu}

\affil*[1]{\orgdiv{Department of Computer Science}, \orgname{University of Illinois Urbana-Champaign}, \orgaddress{\street{201 North Goodwin Avenue}, \city{Urbana}, \postcode{61801}, \state{IL}, \country{USA}}}

\affil[2]{\orgdiv{Department of Computer Science}, \orgname{Utah State University}, \orgaddress{\street{4205 Old Main Hill}, \city{Logan}, \postcode{84322}, \state{UT}, \country{USA}}}


\abstract{In mathematical proof education, there remains a need for interventions 
that help students learn to write mathematical proofs. Research has shown that 
timely feedback can be very helpful to students learning new skills. 
While for many years natural language processing models have struggled to perform 
well on tasks related to mathematical texts, recent developments in natural 
language processing have created the opportunity to complete the task of giving 
students instant feedback on their mathematical proofs. In this paper, we present 
a set of training methods and models capable of autograding freeform mathematical 
proofs by leveraging 
existing large language models and other machine learning 
techniques. The models are trained using proof data collected from four different 
proof by induction problems. We use four different robust large language models to 
compare their performances, and all achieve satisfactory performances to various 
degrees. Additionally, we recruit human graders to grade the same proofs as the 
training data, and find that the best grading model is also more accurate than most human 
graders in terms of grading accuracy. With the development of these grading models, we create and 
deploy an autograder for proof by induction problems and perform a user study with students. 
Results from the study shows that students are able to make significant 
improvements to their proofs using the feedback from the autograder, but students 
still do not trust the AI autograders as much as they trust human graders.
Future work can improve on the autograder feedback and figure out ways to help 
students trust AI autograders.}




\maketitle

\section{Introduction}

Writing mathematical proofs has been identified as an 
important~\cite{acm2013curriculum,acm2016curriculum,acm2014curriculum}
and yet challenging topic~\cite{goldman2008identifying} in computing and 
mathematics education. 
A large body of research has shown that  
timely feedback is crucial to student learning~\cite{anderson1995cognitive,shute2008feedback}. 
However, students are largely unable to receive timely feedback on 
written proofs due to the need to have proofs collected and 
hand-graded by instructors or teaching assistants.
The ability to grade student proofs fully automatically with natural language 
processing (NLP) alleviates this need by allowing us to give students instant 
feedback on their proofs to let students iteratively enhance the quality of their 
proofs.



In this paper, we propose a novel set of training methods and models capable of 
autograding freeform mathematical proofs, a problem at the intersection of 
mathematical proof education and 
Automatic Short Answer Grading (ASAG), by using existing NLP models and other 
machine learning techniques. 
Our proof autograder enables the development of grading systems that provide 
instant feedback to students without needing attention from instructors. It can 
also be deployed in large-scale educational platforms, allowing for more 
access for students. 


The main contributions of this paper are: 
\begin{itemize}
\item Introducing the first pipeline of machine learning models capable of autograding mathematical 
proofs with similar accuracy to human graders 
\item Quantifying the amount of training data needed to achieve a satisfactory 
performance from the grading models
\item Publishing an anonymized and labeled mathematical proof dataset that can be 
used in future model developments~\cite{poulsen2024student}
\item Creating a set of autograded problems using the grading pipeline, and performing a user study that answers the following research questions: 
\begin{itemize}
    \item Are students able to write better proofs by interacting with the 
    autograder and the feedback it generates?
    \item Are students satisfied with the autograder and the feedback it provides?
    \item Does using the autograder make students more willing to use similar AI 
    autograders in the future? 
\end{itemize}
\end{itemize}

The rest of the paper is organized as follows: we first introduce the related work 
in Section~\ref{sec:literature}, and then present the source and preprocessing of 
the training data for our model in~\ref{sec:data}. We then introduce the large 
language models used in the grading process and our own model for grade 
calculation in Sections~\ref{sec:llm} and~\ref{sec:linear}. 
Section~\ref{sec:human} gives the methods for the comparison to human 
graders. In Section~\ref{sec:results}, we show the performances from each large 
language model on the data, and compare the model grading against human grader 
performances. 
In Sections~\ref{sec:user} and~\ref{sec:user_results}, we provide detail about a 
user study where we recruited students to write and improve their proofs using the 
developed autograders, and collected their progress and feedback with the 
autograder. 

\section{Related Work}\label{sec:literature}

\subsection{Research on Mathematical Proof Education}
\new{It has been acknowledged that teaching and using proofs plays an important role in 
mathematics education~\cite{knuth2002secondary}. }
While there has been extensive work both on software tooling and its evaluation 
for helping students learn school geometry proofs~\cite{anderson1995cognitive},
there is still a need for tools to provide students with interventions and 
scaffolding on learning to write college-level mathematical proofs. Stylianides 
and Stylianides' review of the literature on teaching and learning proofs 
concluded that ``more intervention-oriented studies in
the area of proof are sorely needed''\cite{stylianides2017research}. 
There have been interventions that focus on helping students understand the need for
proofs, as students often believe empirical arguments without seeing the need for
proof~\cite{stylianides2017research,jahnke2013understanding,brown2014skepticism,stylianides2009facilitating}. 
There are also various experience reports
on novel interventions that instructors have tried with limited evidence on their
effectiveness~\cite{harel2001development,larsen2008proofs,norton2022addressing}.
Several tools have been created which provide a visual method for students to 
construct simple mathematical 
proofs~\cite{breitner2016incredible,lerner2015polymorphic,seth2022proofblock}. 
While these tools can act as scaffolding for students in the early stages of 
learning to write proofs, there are currently no existing software tools that can 
provide help to students working on the authentic task of writing proofs on their 
own from scratch.

\new{Within the field of mathematical proofs, proof by induction is a particular point of interest 
because of its requirement for various cognitive schemas and abstraction skills~\cite{movshovitz-hadar1993false,dubinsky1986teaching}. 
Its mostly non-explanatory nature~\cite{lange2009why} often leaves students uncertain 
about when to chose induction over other explanatory proof techniques.
Previous research has explored methods for teaching students both when to use induction~\cite{stylianides2016conditions}, 
and how to construct induction proofs correctly~\cite{dubinsky1986teaching,dubinsky1989teaching}. 
Several misconceptions have been identified during the process~\cite{baker1996students,dubinsky1989teaching,norton2022addressing,stylianides2007preservice}. 
In Section~\ref{sec:data}, we will discuss how the items in our rubric align with many of the 
misconceptions reported in the literature. 
A more exhaustive list of all misconceptions can be found in recent work in mathematics 
education~\cite{norton2022addressing,fluckiger2025student}. 
Most of the misconceptions listed in prior work have also shown up in our own training data set~\cite{fluckiger2025student}.
Despite the prevalence of these issues, students currently lack tools to help 
them address these misconceptions and practice constructing induction proofs effectively.}
The grading models introduced in this paper enable the development of software 
tools that provide students with automated feedback on their work as they practice 
writing natural language mathematical proofs on their own.

\subsection{Natural Language Processing (NLP)}
As a subfield of artificial intelligence, NLP primarily focuses on training and 
using models to interpret and generate natural languages. 
In 2017, NLP researchers developed the Transformer architecture, which 
relies solely on attention mechanisms, achieving good performance in learning the 
dependencies in inputs and outputs~\cite{vaswani2023attention}. It also enabled easier training and inference 
by lowering the need of advanced training equipments or extended training 
time~\cite{vaswani2023attention}. Further developments have led to the 
introduction of various 
pretrained large language models accessible such as BERT, Llama and Llama2, and 
OpenAI's GPT 
family of models 
~\cite{brown2020gpt3,devlin2019bert,touvron2023llama}. 
In addition to being able to generate text, these models can take texts as inputs 
and return vectors representing the words or 
sentences in the texts. These vectors are called embeddings. 

The word embedding method is a major part of the current field of NLP aiming to map 
words and phrases into a multi-dimensional continuous vector space. Each vector 
captures the semantic meaning of each word and phrase, and researchers can use the 
vectors to perform downstream tasks such as word prediction and sentence 
classficition~\cite{haller2022NLP}. Multiple Word embeddings of a sentence can 
also be combined into one sentence embedding through the use of an extra network 
to capture the meaning of the whole sentence~\cite{kiros2015skip-thought}. In 
essence, sentence embeddings capture the semantic meaning of the whole sentence. 

\subsection{Automatic Short Answer Grading (ASAG)}
Researchers have attempted to use sentence embeddings to perform ASAG. Various 
NLP models have been tested on their performance, with attention-based models 
performing relatively well in terms of accuracy, latency, reproducibility, and 
ease in fine-tuning~\cite{bonthu2021automated}. Existing training data sets cover 
topics in data structures, introductory statistics, and a few other topics, with 
no coverage of mathematical proof problems~\cite{bonthu2021automated}. 
\new{One thread of previous work utilized BERT in automatic scoring of reading comprehension 
by fine-tuning BERT with a set of reading comprehension problems and answers, 
achieving accuracy almost as good as human graders~\cite{fernandez2023automated}. 
However, it lacks the ability to easily transfer to new problems, creating the need 
to fine-tune the model repeatedly when new problems are added, which can be a computationally expensive process. 
The same drawback would apply to other ASAG attempts utilizing fine-tuning as the primary way of grading. }
Previous work has also produced systems 
able to grade student's descriptions of short snippets of Python code by creating item-specific models, achieving 
over 85\% accuracy with a dataset of less then 600 samples~\cite{binglin2022eipe}. 
These models have successfully been deployed in 
the classroom to help students receive timely feedback and reduce instructor 
grading load~\cite{azad2020strategies}.

Efforts to apply ASAG in mathematical contexts have been made but have achieved only limited success to date.
Researchers have developed Mathematical Language Processing to perform mathematical ASAG, and achieved an absolute mean error of $0.04$ out of a full grade of $3$~\cite{lan2015symbolic}. 
However, the approach taken by Mathematical Language Processing allows only a 
limited set of mathematical 
vocabulary, and relies heavily on symbolic manipulation on the answers. As a 
result, it cannot autograde mathematical proofs, which usually 
involves reasoning using natural language, such as proof by induction or proof on 
graphs. 
MathBERT was created by fine-tuning BERT, aimed at dealing with a larger 
mathematical vocabulary and more complex mathematical 
contexts~\cite{shen2023mathbert}. It achieved over $90\%$ accuracy on autograding 
tasks. However, the problems addressed were one-sentence Q\&A tasks that required 
minimum reasoning or formula manipulation. These tasks were significantly shorter than the full 
inductive proofs considered in this work. To deal with verifying longer proofs, 
researchers have also attempted autoformalization~\cite{wu2022autoformalization}, 
translating natural language into formal logic. This should make checking the 
correctness of the steps in a proof easier. However, even with several years of 
research, none of the existing work is able to translate even half of the proofs 
to formal logic~\cite{wu2022autoformalization}. The low success rate makes 
autoformalization not suited for grading mathematical proofs at the moment. 

To our knowledge, no tools have successfully dealt with grading mathematical 
proofs using natural language processing. More emphasis needs to be put on 
verifying the natural language reasoning 
part of the proofs in addition to the arithmetic formulae. 

Utilizing ASAG in practice poses a few difficulties. 
First of all, it is hard for the ASAG systems to be perfect: mistakes will likely happen during the grading process. 
In the context of potentially inaccurate feedback from the graders, helping students navigate through the feedback can be crucial in achieving the full potential from ASAG systems~\cite{li2023wrong}. 
Another challenge facing the wider use of ASAG is students' trust in AI. 
Previous research around AI trust has shown that people are more likely to trust human decisions than algorithmic decisions, 
especially when the tasks are more subjective~\cite{lee2018understanding,castelo2019task-dependent}. 
Some previous work has observed that students underestimated the accuracy of some ASAG systems, believing they were graded incorrectly when the grades were accurate~\cite{hsu2021attitudes}. 
The lack of trust in the technology behind ASAG systems could hinder their effectiveness. 

\section{Model Development}

We propose an automated grading process with two steps. First, we run student-written proofs 
through a pretrained large language model to get the embeddings. We then apply our custom grading 
model to the embeddings to calculate accuracy on various rubric points. We also 
recruited human graders to re-grade proofs in our data set, which are used for baseline 
comparison.

\subsection{Data}\label{sec:data}
The proofs used in this study were collected from the pretest and posttest of a series of studies designed 
to measure the learning gains of students using Proof 
Blocks~\cite{poulsen2023efficiency,poulsen2023measuring,poulsen2024disentangling}.
The participants were students recruited from a Discrete Mathematics course 
at the University of Illinois Urbana-Champaign. The proofs were originally 
collected in the form of 
markdown files, and later graded by members of the 
research team. 
The proofs were graded in a multi-step process which involved multiple members of 
the research team grading the proofs independently, and then meeting together to 
discuss disagreements. The graders were able to reach an inter-rater reliability 
rating of Cronbach's $\alpha$ at 0.82, 0.88, and 0.92 for each of the three 
studies 
respectively~\cite{poulsen2023efficiency,poulsen2023measuring,poulsen2024disentangling}.

These 
measurements show very high consistency in agreement between researchers,
allowing us to use these labels as ground truth for training and further analysis.
Since the grading of informally written mathematical proofs can be subjective, we 
recognize that our AI autograders and the human baseline comparison must not only 
agree with fundamental truths but also align with the grading standards set by the 
research team that created the original labels. However, we note that this is the 
same task that an ASAG system or a teaching assistant must accomplish when grading 
student work for a course—not only verifying underlying truth but also adhering to 
the grading criteria set by the instructor, who acts as the ground truth in this 
context.
We also believe that our use of only technical, and not stylistic rubric 
points, 
as well as the rigorous process we used to have multiple researchers agree on a 
grading scheme (rather than have them decided by a single instructor), make our 
grading system as objective as possible in this context.

The proofs were graded using seven rubric points, each identifying a step of the 
induction proof. 

\begin{itemize}[leftmargin=.5in]
    \item[\textbf{R1}] Identifying the base case(s)
    \item[\textbf{R2}] Proving the base case(s)
    \item[\textbf{R3}] Stating the inductive hypothesis
    \item[\textbf{R4}] Setting the bound of the inductive hypothesis
    \item[\textbf{R5}] Stating the goal of the inductive step
    \item[\textbf{R6}] Breaking down the inductive step
    \item[\textbf{R7}] Applying the inductive hypothesis
\end{itemize}

\new{Each of these rubric points was designed to address specific aspects of constructing a 
complete and correct induction proof. The rubric points also align with common misconceptions 
identified in prior research.  
Students' difficulties in identifying and proving base cases (R1 and R2) were discussed 
by~\citet{norton2022addressing}, who also reference earlier literature documenting these 
issues.
Many students have difficulties in stating the inductive assumption correctly. For example, some 
students will assume the entire conjecture to be true rather than the inductive 
implication~\cite{norton2022addressing,movshovitz-hadar1993false}. Students will also fail to 
give the correct upper or lower bound for the inductive implication~\cite{fluckiger2025student}. 
These issues are captured by rubric points R3 and R4.
The rubric points R5, R6, and R7 correspond to the misconception ``Sets up induction proof 
properly 
but does not succeed 
in proving that $P(n) \Rightarrow P(n+1)$" documented by Dubinsky~\cite{dubinsky1989teaching}.
The grading instructions included in our public dataset~\cite{poulsen2024student} give detailed 
instructions on when each rubric point should be marked as correct for a given proof, including 
examples.
}

We selected four different proof by induction problems, and labeled them as P1, 
P2, P3, P4. 
P1 focuses on induction with a recursively defined function, and has 1623 
non-empty proofs collected; P2 and P3 are induction proving the closed form 
solution to a summation formula, and have 1288 and 342 proofs respectively; P4 
proves the divisibility of a function over the natural numbers, and has 333 
proofs. The problems statements 
are shown in Table~\ref{tab:problem-statements}. 
\begin{table}[h]
\setlength{\tabcolsep}{0.5em}
\renewcommand{\arraystretch}{1.5}

\begin{tabular}{|c|p{4.3in}|}
\hline
~P1~ & 
    Suppose that $g: \mathbb{N} \rightarrow \mathbb{R}$ is defined by
$g(0) = 0$, 
$g(1) = \frac{4}{3}$, and
$g(n) = \frac{4}{3}g(n-1) - \frac{1}{3}g(n-2)$ for $n \geq 2$.
Use induction to prove that $g(n) = 2 - \frac{2}{3^n}$ for any natural 
number $n$.
\\ \hline
~P2~ & 
Prove the following statement by induction.
$\forall n \in \mathbb{N}, \sum_{i=0}^n i = \frac{n(n+1)}{2}$.
\\ \hline
~P3~ & 
Use (strong) induction to prove the following claim:
    
$\sum_{p=0}^n (p \cdot p!) = (n+1)! - 1$
for any natural number $n$. 
\\ \hline
~P4~ & 
Use (strong) induction to prove the following claim: 

for any natural number $n$, $2n^3+3n^2+n$ is divisible by $6$. 
\\ \hline
\end{tabular}
\caption{Problem statements for P1-P4.}
\label{tab:problem-statements}
\end{table}
The proofs were initially graded as $0$, $1$, $2$ for each rubric point. A 
grade of $0$ is given if the required rubric point is not present, $1$ if the 
rubric point is partially correct, and $2$ if it is completely correct. 
In our work, we collapse the grades into two categories, correct and incorrect. 
Original labels of $0$ and $1$ are combined into $0$, standing for incorrect 
proofs; original labels of $2$ are changed to $1$ to represent fully correct 
proofs. This allows us to give students individualized feedback on the correctness 
of each of the seven rubric points. Future work may use more complex labels for 
even more granular feedback.
One sample proof and its R1-R7 labels are shown in Figure~\ref{fig:sample-proof}. 
\begin{figure}
\begin{framed}
\raggedright
FOr all natural numbers n; let the proposition P(n) be $\sum_{i=0}^{i}$ (from 0 to 
n) = to n(n+1)/2

P(1) is $\sum_{i=0}^{i}$ (sigma from 0 to 1) = to 1(1+1)/2
Left hand side =1; right hand side = 1
So P(1) is true

Therefore let P(x) be, P(x) = $\sum_{i=0}^{i}$ (sigma from 0 to x) = x(x+1)/2

Therefore assume P(x+1) = $\sum_{i=0}^{i}$ (sigma from 0 to x+1) = (x+1)(x+2)/2

the difference of sums between sigma to x and sigma to x+1 is the term x +1 only if we subtract P(x) from P(x+1) the reamaing number sould be x+1

(x+1)(x+2)/2 - x(x+1)/2 =  $\frac{x^2 + x}{2}-\frac{x^2+3x+2}{2}$ = $ \frac{2x +2}{2}$ = x + 1

Therefore its proved by mathematical induction
\end{framed}
\vspace{1mm}
\caption{An example proof for P2.
In the original labeling by previous researchers, this proof was labeled correct 
on rubric points R1 (Identifying 
the base case), R2 (Proving the base case), and R6 (Breaking down the inductive 
step), 
and incorrect for all other rubric points. 
}
\label{fig:sample-proof}
\end{figure}
We have published our data set in a public data repository so that future 
researchers may replicate our work and make further improvements in autograding 
mathematical proofs~\cite{poulsen2024student}.

\subsection{Model selection for Embeddings}\label{sec:llm}
Several pretrained large language models are capable of calculating embeddings 
such as BERT and its mathematically fine-tuned version 
MathBERT~\cite{shen2023mathbert}, Llama2 and its mathematically fine-tuned version 
Llemma~\cite{azerbayev2023llemma}, and GPT-3~\cite{openaiembeddingsdocs}. Base 
BERT and base Llama2 are excluded because MathBERT and Llemma have been observed 
to perform better than their corresponding base models when dealing with 
mathematical contexts~\cite{shen2023mathbert,azerbayev2023llemma}. 
MathBERT is among the best pretrained language models at mathematical language, 
and runs on a CPU. The memory requirement is typically less than 600MB. Because of 
the low memory requirement, MathBERT allows for cheap deployment. 
It is worthwhile trying MathBERT despite having fewer parameters than most large 
language models, because if MathBERT can achieve similar performances as the other 
large language models, it would be cheaper to deploy. This could become a critical factor 
when selecting the grading models if they need to be scaled in the future to accommodate a larger number of students.
GPT-3 is the best accessible language model at the moment for embeddings. We use the latest embedding endpoint text-embedding-3-large for analysis, as it outperforms the old endpoint text-embedding-ada-002 by 0.3\% in accuracy in our grading tasks. GPT-3
requires API calls 
for embeddings, and each API call costs a small amount of money. This additional cost should be taken into consideration when scaling this grading model for larger number of students. We did not use 
GPT-4 as it does not yet support embeddings, which are preferred to text 
completion for classification tasks such as ours~\cite{openaiembeddingsdocs}.
Llemma has two versions, Llemma7b with 7 billion parameters, and Llemma34b with 34 
billion parameters. Both can be run locally, but take a significant amount of RAM 
and require a GPU even for inference. 
In our study, we utilize MathBERT, GPT-3, Llemma34b, and Llemma7b for embeddings and then compare the performances of the models. 

One extra step of data manipulation was performed only for the MathBERT model. 
Currently, MathBERT is unable to recognize certain mathematical or mathematical 
LaTeX notation 
such as ``\verb|f(x)|'' or ``\verb|\sum_{i=1}^n|''. 
\new{These symbols would be 
recognized as individual characters in MathBERT, introducing unnecessary noise 
and lowering its ability to fully understand the contents. 
We modified the tokens from MathBERT so that it would recognize function calls and 
some latex expressions in the same token, enhancing the model's ability to 
understand the words and phrases accurately.}
The modification ensured that every proof in the dataset fit 
within the input token limit. \new{Incorporating these combined tokens led to a slight improvement in the model's grading accuracy. For this reason, our analyses were based exclusively on the modified MathBERT. 
}

\subsection{Model fitting}\label{sec:linear}
The downstream task is to grade the proofs using the embeddings as inputs. 
Because the LLM we use for embedding handles the difficult task of extracting the 
relevant features in the text, we were able to use a relatively simple model for 
classification afterwards. 
We tried both linear regression and SVMs, with similar results, and so we chose to 
use linear regression for its simplicity. More complexity can be added if needed 
in the future.
We trained one grading model for each rubric point of each problem, as the exact 
expectations of the rubric points might not be generalizable 
across different problems. 

The input dimension of the grading model is the same as the returned sentence 
embedding dimension of the pretrained model. By default, MathBERT embeddings have 
length $768$, GPT-3 embeddings have length $1536$, and Llemma embeddings have 
$4096$ and $8192$ entries respectively for the two versions. The output dimension is 
set to $2$ to match the two distinct categories for grading: ``incorrect'' (0) and 
``correct'' (1). The model produces a probability score from the input embeddings, and 
normalizes using the softmax function to yield the probability distribution over 
the two classes. We will use this probability distribution to test the model and to autograde future submissions. 

The dataset is broken down into training, testing, and validation sets, with 
$15\%$ for testing and another $15\%$ for validation. The remaining $70\%$ are 
used for training. Performance on the test set will be used to compare the 
performances of models. The $15\%$ data for validation is reserved for 
future work with increased model complexity. 

For the training of our grading model, we chose a batch size of $128$ as it is a common 
choice for training networks. Number of epochs for training range from $100$ to 
$1000$, with a step size of $100$ epochs. Initially, the learning rate is set to 
$0$ and linearly increased to a peak value of $0.001$ over the first $60\%$ of the 
training epochs. This linear increase provides a warm-up period for the model to 
gradually adapt to the task. Subsequently, during the second half of the training, 
the learning rate is exponentially decreased to $\frac{1}{10}$ of the peak value, 
allowing the model to fine-tune its parameters and converge towards a stable 
solution.

To save computational resources, we generate all embeddings by running the pretrained models once on the 
entire proof dataset. Then we save the mapping between each proof and its corresponding embedding to a pickle file. We ran the two 
Llemma models on our University's supercomputing cluster, using eight NVIDIA 
A100 GPUs simultaneously, taking about one second on average to turn one proof into embeddings; GPT-3 embeddings were retrieved from an OpenAI API 
call, and we ran MathBERT on a single NVIDIA RTX4090 GPU. The customized grading 
models were built and trained using PyTorch~\cite{adam2019torch} on the same NVIDIA 
GPU. For each model, trainings completed at a rate of about 2 seconds per 100 
training epochs and 10 batchs. 

\subsection{Comparison to human graders}\label{sec:human}
Theoretically, it is hard for machine learning models to perfectly grade the 
freeform mathematical proofs. However, it can also be hard for humans to agree on 
the grading of mathematical proofs both due to ambiguity in language and honest 
mistakes. In the case that there is any subjectivity in the grading, human graders 
also need to match the specification set by the course instructor.
To our knowledge, no 
prior study has quantified the accuracy of humans grading mathematical 
proofs to match an instructor specification. Additionoally, we want to compare how 
closely human graders can match the grading specification relative to our NLP 
grading models. 

Using the last author's professional network, we recruited graduate students from 
several universities who had been teaching assistants in 
discrete mathematics courses. 
These graduate students all had experience grading mathematical induction proofs.
A total of 9 graduate students from 4 different institutions participated in the grading tasks. 
For each human grader, we sent 15 student proofs from each of the 4 proof problems 
(for a total of 60 problems 
each), and asked them to grade these proofs using the same rubric. 
Along with the proofs to grade, we also provided 15 examples of graded proofs for each 
of the problems, the original problem statements, and detailed notes with 
explanations of the rubrics created by previous researchers. This combination of 
example labeled proofs and detailed rubric explanations were designed to set the 
graders up to be as successful as possible on their grading task. In fact, we 
provided the graders with more training materials for grading than many 
instructors 
do for grading of actual courses (in our experience).
The graders were also free to ask us any questions regarding problems encountered 
during the grading process just as they would be when grading student work in a real course setting. Each grader 
received a \$100 gift card as compensation for their grading efforts. This human 
subjects data collection procedure was approved by the institutional review board 
at Utah State University. 

\section{Model Performance Results}\label{sec:results}

In this section, we provide an overview of the grading models' performance, the accuracy of the human graders, and an analysis of the amount of training data required to get satisfactory performance. 

\subsection{Grading Model Performance}
For each model and each training epoch, we calculate the performance using the testing set, comparing the model predictions to ground truth labels that previous researchers created. 
We compute the confusion matrix to obtain the number of true positive, true 
negative, false positive, false negative cases, and then we calculate the accuracies 
and F1 scores. Among the models using different training epochs, we select the one 
with the highest accuracy. 
\new{Rubric point accuracy metrics for each problem and each pretrained model are summarized by 
averaging the accuracies from the testing set for all rubric points. 
RMSE and Pearson's $r$ are calculated by adding up the results for each rubric point, 
and normalizing the grades to between 0 and 100.} 
The results are shown in Table~\ref{tab:model_results_overview}. 

\begin{table}[h]
\centering
\begin{tabular}{ c  c  c  c  c  c } 

Problem & Data Size & Grading Method & Accuracy & \new{RMSE} & \new{Pearson's $r$} \\ \toprule 

\multirow{5}{*}{P1} & \multirow{5}{*}{1623} & MathBERT	\& linear & 83.6\% & 23.35 & 0.79 \\ \cmidrule(l){3-6}
& & GPT-3 \& linear & 87.9\% & 17.06 & 0.90 \\ \cmidrule(l){3-6}
& & Llemma7b \& linear & 90.7\% & 14.53 & 0.92 \\ \cmidrule(l){3-6}
& & Llemma34b \& linear & 90.2\% & 15.33 & 0.92 \\ \cmidrule(l){3-6}
& & Human Graders & 84.1\% & 22.20 & 0.81 \\ \cmidrule(l){1-6}

\multirow{5}{*}{P2} & \multirow{5}{*}{1288} & MathBERT\&linear & 83.7\% & 27.88 & 0.77 \\ \cmidrule(lr){3-6}
& & GPT-3 \& linear & 88.3\% & 22.32 & 0.86 \\ \cmidrule(lr){3-6}
& & Llemma7b \& linear & 90.9\% & 16.28 & 0.92 \\ \cmidrule(lr){3-6}
& & Llemma34b \& linear & 90.7\% & 17.87 & 0.91 \\ \cmidrule(lr){3-6}
& & Human Graders & 91.3\% & 13.58 & 0.94 \\ \cmidrule(l){1-6}

\multirow{5}{*}{P3} & \multirow{5}{*}{342} & MathBERT\&linear & 85.7\% & 20.51 & 0.85 \\ \cmidrule(lr){3-6}
& & GPT-3 \& linear & 89.2\% & 17.07 & 0.88 \\ \cmidrule(lr){3-6}
& & Llemma7b \& linear & 85.7\% & 21.98 & 0.81 \\ \cmidrule(lr){3-6}
& & Llemma34b \& linear & 86.6\% & 18.59 & 0.87 \\ \cmidrule(lr){3-6}
& & Human Graders & 85.1\% & 23.55 & 0.78 \\ \cmidrule(l){1-6}

\multirow{5}{*}{P4} & \multirow{5}{*}{333} & MathBERT\&linear & 86.3\% & 21.13 & 0.76 \\ \cmidrule(lr){3-6}
& & GPT-3 \& linear & 85.4\% & 17.86 & 0.83 \\ \cmidrule(lr){3-6}
& & Llemma7b \& linear & 89.9\% & 13.68 & 0.91 \\ \cmidrule(lr){3-6}
& & Llemma34b \& linear & 90.2\% & 13.98 & 0.91 \\ \cmidrule(lr){3-6}
& & Human Graders & 85.8\% & 25.56 & 0.71 \\ 

\end{tabular}
    \caption{Overall grading accuracy for grading models and human graders. Llemma-based grading models have the highest accuracy on average, and slightly higher than human graders. }
    \label{tab:model_results_overview}
\end{table}

\new{All four pretrained models achieve over $80\%$ average accuracy, and achieve higher RMSE and Pearson's $r$ metrics than most of the earlier ASAG work~\cite{bonthu2021automated}.} 
MathBERT is the lowest performing one with an average of $84.1\%$ accuracy across all problems, as it is the model with the fewest parameters, thus capturing fewer details accurately. 
The best grading models are Llemma-based, both Llemma34b and Llemma7b, with $90.0\%$ and $90.2\%$ average accuracy, even 
beating GPT-3-based models. This is likely because Llemma is more specifically 
trained to handle mathematical language. Moreover, the large number of parameters 
gives Llemma the ability to capture almost the same amount of information as 
GPT-3. 
The accuracies and F1 scores of Llemma34b graders on a by-rubric-point basis appears in Table~\ref{tab:model_results_best}. 

\begin{table}[h]
    \centering
        \begin{tabular}{| c | C{.43cm} | C{.43cm} | C{.43cm} | C{.43cm} | C{.43cm} | C{.43cm} | C{.43cm} | C{.43cm} | C{.43cm} | C{.43cm} | C{.43cm} | C{.43cm} | C{.43cm} | C{.43cm} |} 
            \hline
             & \multicolumn{2}{c|}{R1} & \multicolumn{2}{c|}{R2} & 
             \multicolumn{2}{c|}{R3} & \multicolumn{2}{c|}{R4} & 
             \multicolumn{2}{c|}{R5} & \multicolumn{2}{c|}{R6} & 
             \multicolumn{2}{c|}{R7} \\ \hline

            & A & F1 & A & F1 & A & F1 & A & F1 & A & F1 & A & F1 & A & F1 \\ \hline

            P1 
            & .888 & .863
            & .862 & .848
            & .892 & .834
            & .897 & .810
            & .909 & .855
            & .931 & .886
            & .935 & .876 \\ \hline
            P2 
            & .935 & .940
            & .929 & .933
            & .891 & .855
            & .875 & .816
            & .875 & .839
            & .918 & .885
            & .924 & .889 \\ \hline
            P3 
            & .878 & .923
            & .816 & .870
            & .878 & .914
            & .939 & .954
            & .694 & .706
            & .918 & .846
            & .939 & .880 \\ \hline
            P4 
            & .979 & .988
            & .938 & .962
            & .917 & .926
            & .896 & .906
            & .812 & .800
            & .896 & .783
            & .875 & .750 \\ \hline

        \end{tabular}
    \caption{Llemma34b model grading performances for all four problems (P1-P4) and all seven rubric points (R1-R7). Column label A stands for accuracy; column label F1 stands for F1 scores. }
    \label{tab:model_results_best}
\end{table}

The results indicate no clear pattern regarding which rubric points can consistently be graded more accurately than others. 
Currently, Llemma and Llemma-based grading models perform well enough for use by 
students as practice for writing mathematical proofs. 
In future work, we want to investigate how to provide more accurate and detailed feedback to students, as the correctness of the models is not guaranteed.

\subsection{Comparison to human graders} \label{sec:human_results}
After the human graders finish the grading process, we calculate the accuracies 
and F1 scores of human grader results and compare them with those of the grading models. 
The overall performance is also shown in Table~\ref{tab:model_results_overview}, 
with the detailed accuracy and F1 score for each rubric point shown in 
Table~\ref{tab:model_results_human}. On average, human graders achieve $86.6\%$ 
agreement with the research team labels, which is higher than MathBERT but lower 
than GPT-3. The comparison provides a solid ground for further developing and 
utilizing our grading models in 
real courses. 

\begin{table}[h]
    \centering
        \begin{tabular}{| c | C{.43cm} | C{.43cm} | C{.43cm} | C{.43cm} | C{.43cm} | C{.43cm} | C{.43cm} | C{.43cm} | C{.43cm} | C{.43cm} | C{.43cm} | C{.43cm} | C{.43cm} | C{.43cm} |} 
            \hline
             & \multicolumn{2}{c|}{R1} & \multicolumn{2}{c|}{R2} & 
             \multicolumn{2}{c|}{R3} & \multicolumn{2}{c|}{R4} & 
             \multicolumn{2}{c|}{R5} & \multicolumn{2}{c|}{R6} & 
             \multicolumn{2}{c|}{R7} \\ \hline

            & A & F1 & A & F1 & A & F1 & A & F1 & A & F1 & A & F1 & A & F1 \\ \hline

            P1 
            & .830 & .810
            & .741 & .745
            & .852 & .815
            & .859 & .808
            & .830 & .763
            & .896 & .860
            & .881 & .800 \\ \hline
            P2 
            & .970 & .970
            & .933 & .926
            & .881 & .846
            & .881 & .818
            & .904 & .876
            & .889 & .839
            & .933 & .903 \\ \hline
            P3 
            & .881 & .919
            & .807 & .852
            & .844 & .871
            & .822 & .838
            & .793 & .741
            & .896 & .767
            & .911 & .786 \\ \hline
            P4 
            & .881 & .921
            & .874 & .912
            & .859 & .882
            & .815 & .834
            & .807 & .750
            & .867 & .640
            & .904 & .698 \\ \hline

        \end{tabular}
    \caption{Human grading accuracy and F1 scores, broken down by rubric point. The data size is different from the model accuracy testing, as each human grader graded 60 proofs in total. }
    \label{tab:model_results_human}
\end{table}

\begin{figure}[h]
    \centering
        \includegraphics[width=\linewidth]{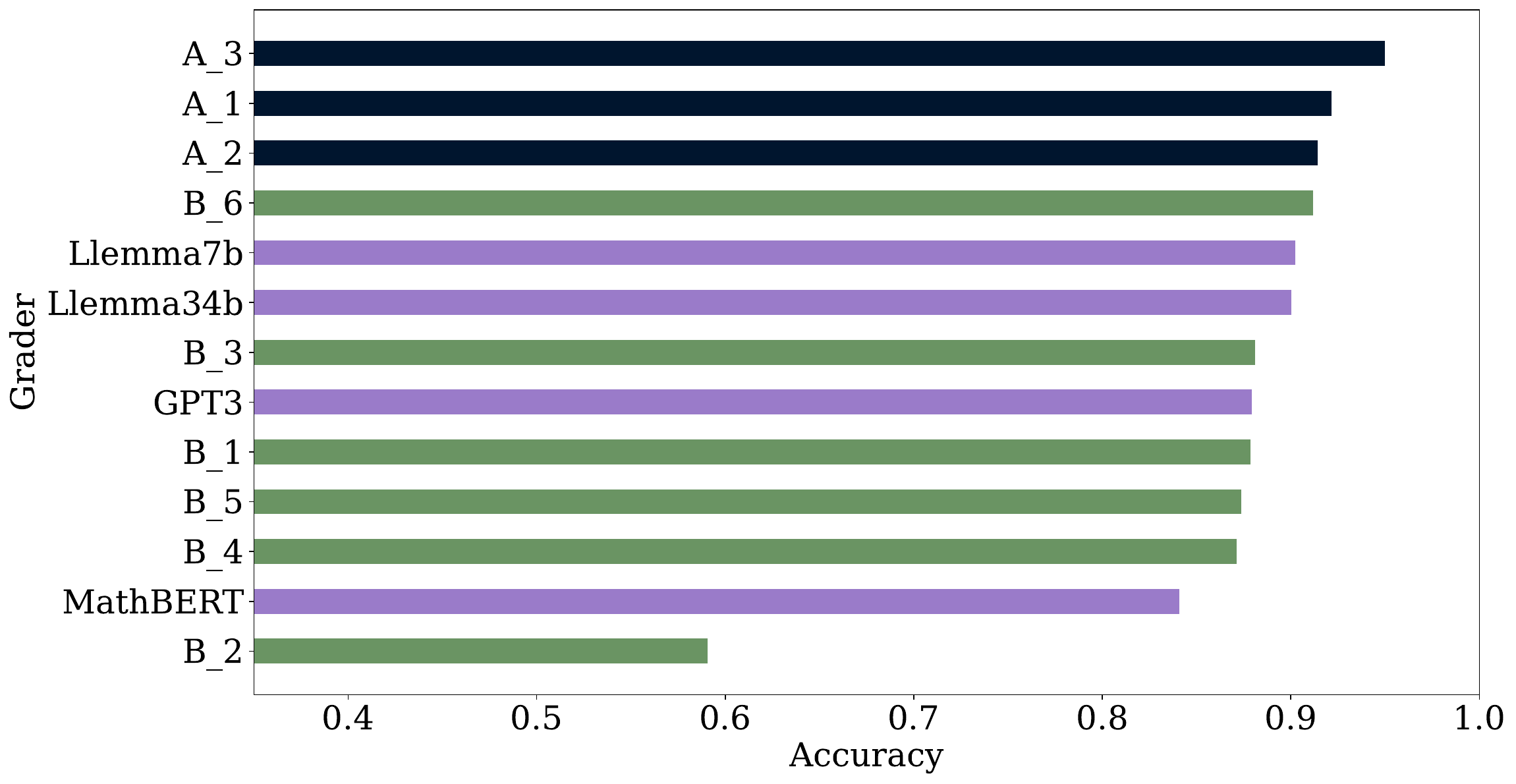}
        \caption{Accuracy for each grading model and human grader. A\_i are the graders from the same university and are more familiar with the used grading rubrics, and B\_i are from other universities. Graders with more experience with the grading rubrics achieve higher accuracy than unexperienced graders and all grading models. }
        \label{fig:per_grader}
\end{figure}

Accuracies for all grading models and individual human graders are plotted in 
Figure~\ref{fig:per_grader}. Human graders are separated into two groups based on whether 
they have been a grader at the same university where the data was collected. Group 
A graders are from the same 
university that the proof data was originally collected at, and thus have more 
experience using the same rubric that was used for the study. Group B graders are 
from other universities, but still have experience in grading proof by 
induction problems (perhaps with different rubrics). The figure shows that in 
general, group A graders are more accurate than group B graders, and are also more 
accurate than the grading models. 
We conclude that our grading models are able to regularly outperform minimally 
trained human graders at matching a grading specification, but human graders with 
more extensive training will be able to outperform our grading models.
This difference is similar to the comparison between graders with different amounts of training in a prior work~\cite{fowler2021eipe}. 

\new{We randomly sampled 20 model-graded submissions and 15 human-graded submissions to P1 and P2 to qualitatively analyze the error patterns of the grading models and human graders. 
One key observation for grading models was the reduced accuracy when handling mathematical formulae. 
Some calculation mistakes went unnoticed by the grading model (for example, writing n=1 as the base case when it should be n=0), 
and some correct values were marked as incorrect (for example, correct base cases of n=0 and n=1 were identified as incorrect by the grader).
On the other hand, human graders were able to grade the formulae correctly, 
but they sometimes ignore subtle details in the proof. For example, 
a correctly but implicitly written bound for the induction hypothesis was marked as incorrect 
by one of the recruited human graders. In addition, TA graders seem to have a hard time 
telling the difference between rubric points R3 (stating the inductive hypothesis) 
and R4 (setting the bound of the inductive hypothesis), marking them both wrong 
when only one of them was actually wrong. The grading model was more effective at making the distinction correctly}

\subsection{How large of a dataset do we need?}
\begin{figure}
\captionsetup[subfigure]{justification=Centering}

\begin{subfigure}[t]{0.49\textwidth}
    \includegraphics[width=\textwidth]{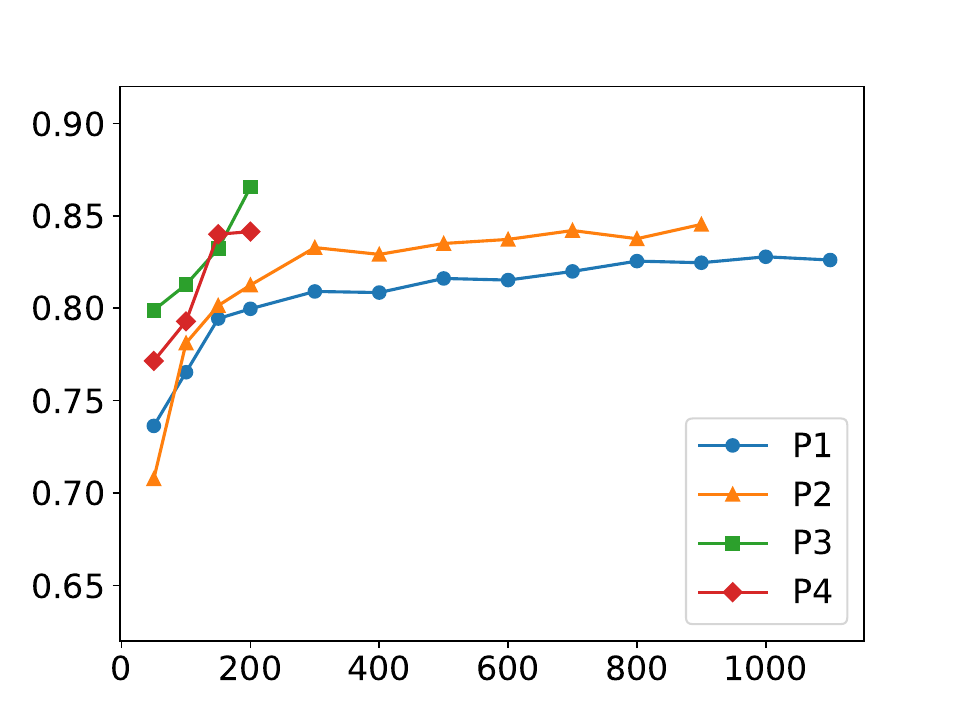}
    \vspace{-2em}
    \caption{MathBERT}
\end{subfigure}\hspace{\fill} 
\begin{subfigure}[t]{0.49\textwidth}
    \includegraphics[width=\linewidth]{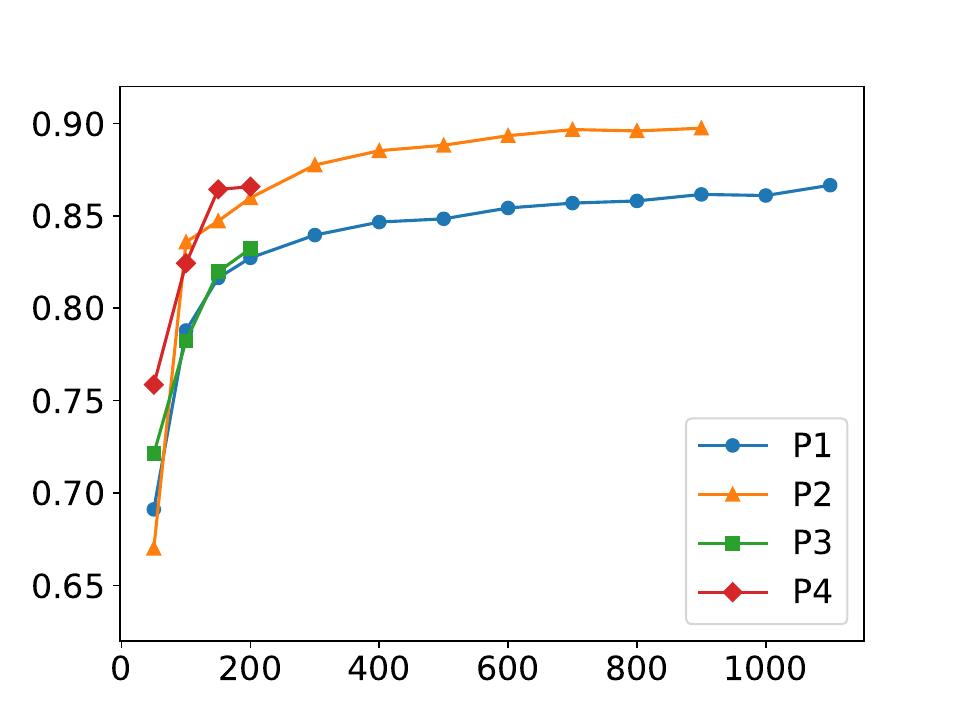}
    \vspace{-2em}
    \caption{GPT-3}
\end{subfigure}

\vspace{-1em}

\begin{subfigure}[t]{0.49\textwidth}
    \includegraphics[width=\linewidth]{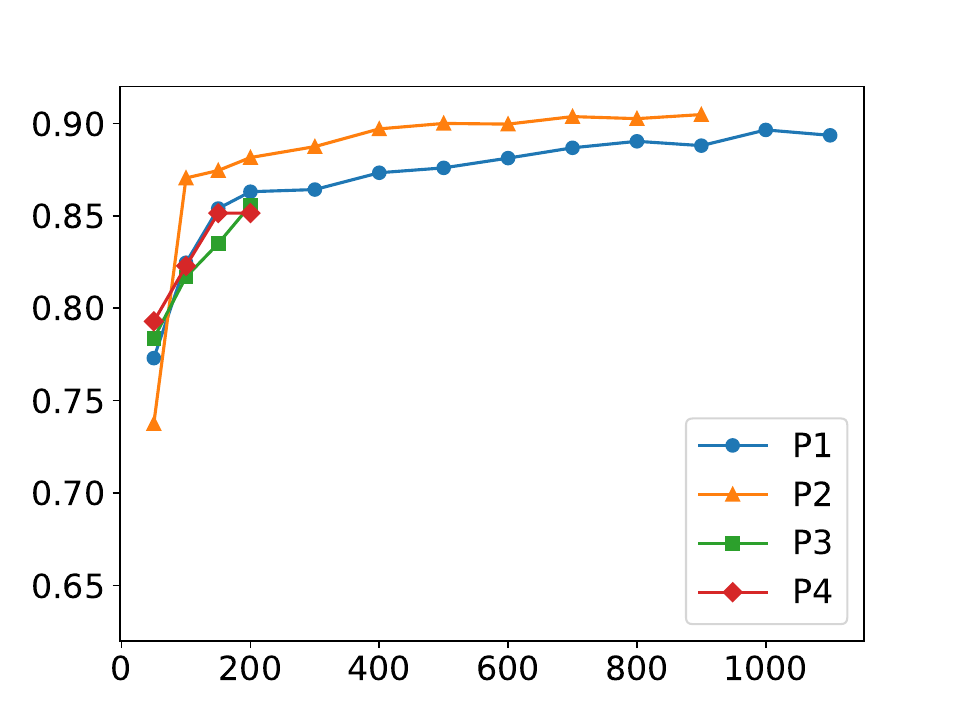}
    \vspace{-2em}
    \caption{Llemma7b}
\end{subfigure}\hspace{\fill} 
\begin{subfigure}[t]{0.49\textwidth}
    \includegraphics[width=\linewidth]{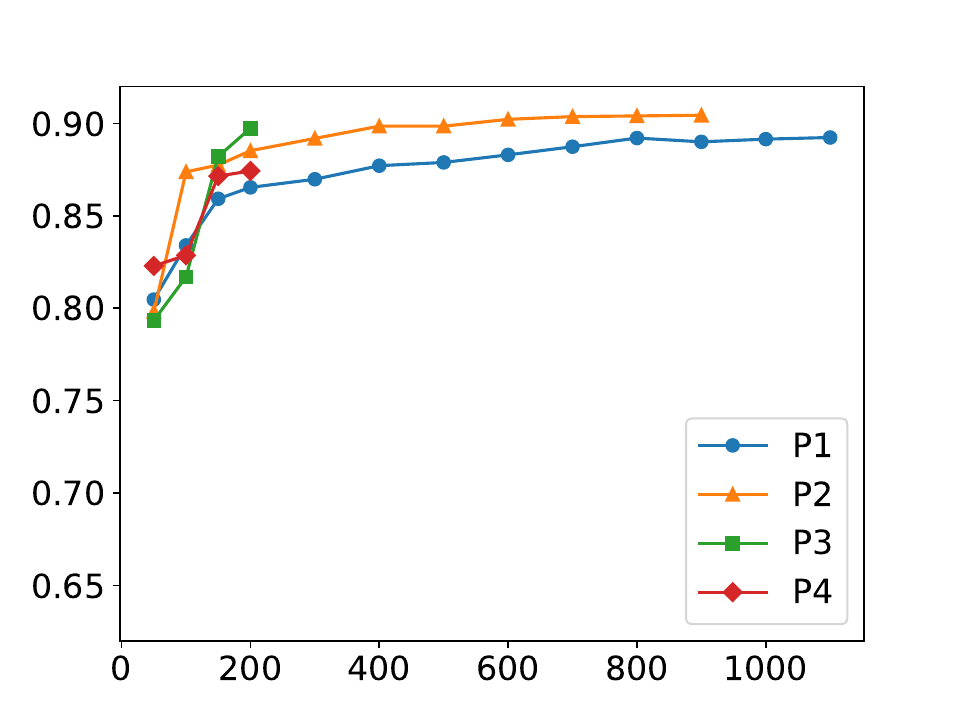}
    \vspace{-2em}
    \caption{Llemma34b}
\end{subfigure}
\vspace{-0.5em}
\caption{Training Size vs. Accuracy. With 400 training data, all grading models are achieving near convergent accuracy, and this accuracy is about 1.2\% away from the maximum accuracy for the grading models. }
\label{fig:size_comp}
\end{figure}

We are also interested in learning how much data is needed to achieve an acceptable performance. We did another set of training and testing to 
compare the 
effect of training data size on model performance. To ensure consistency, the same 
$30\%$ of the data is used as the testing set. The training set sizes range from 
$50$ to the max available size, with a step size of $50$ (before size 200) or 
$100$ (after size 200). The model performance on each training data size is shown 
in Figure~\ref{fig:size_comp}. 

For all four models, the performance is still increasing even with 1100 training data points, so there is space for improvement by collecting more data. Model performance might also be better by making the grading models more complex, such as adding extra layers. 
However, most of the improvement is achieved with the 400 training size threshold. For Llemma34b, the performance at 400 training data already achieves 98.8\% of the full data size accuracy. Based on these results, we conclude that training a model with 400 proofs is sufficient for adequate results. 

\section{User Study Methods} \label{sec:user}
\begin{figure}[h!]
    \centering
    \captionsetup[subfigure]{justification=Centering}
    \begin{subfigure}[t]{.9\textwidth}
        \includegraphics[width=\linewidth]{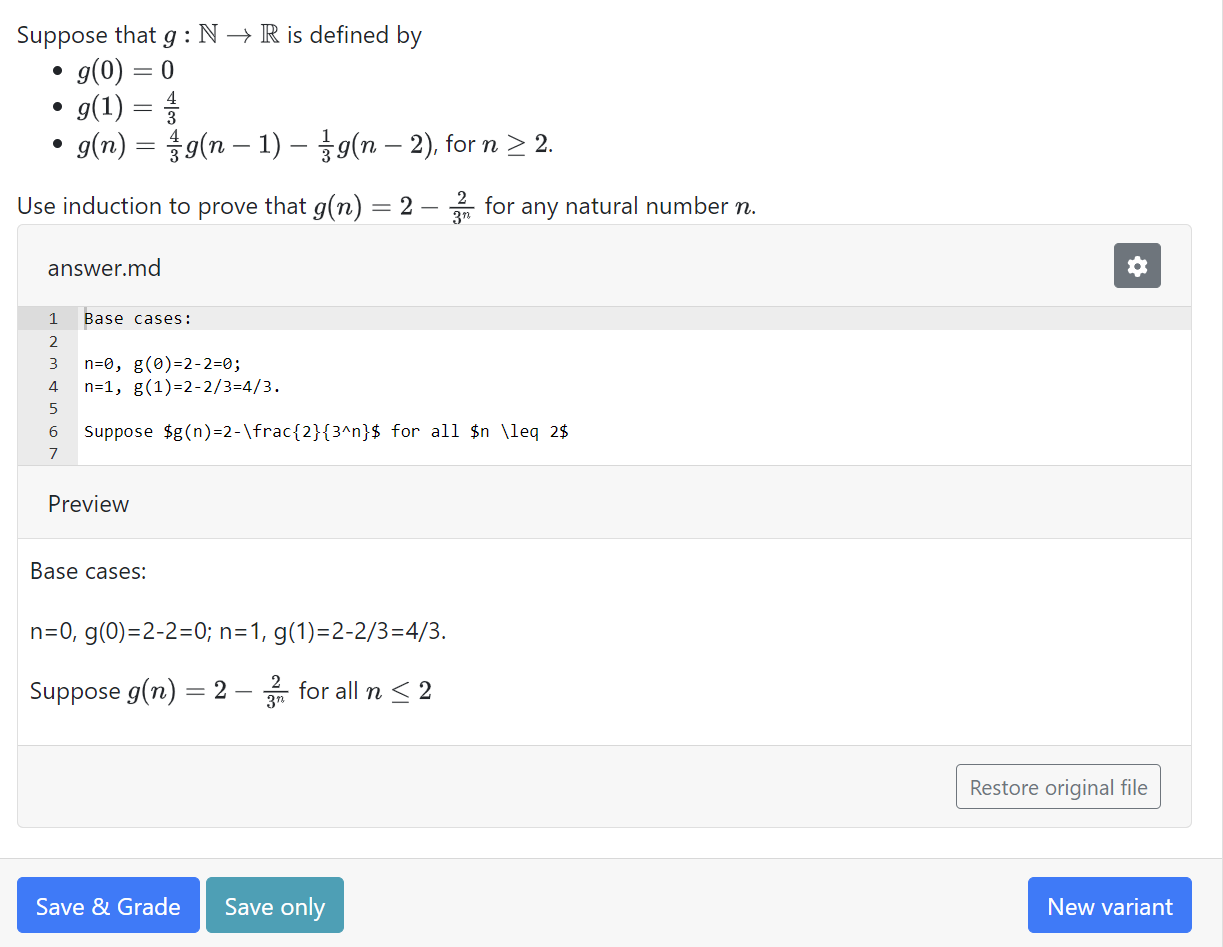}
        \caption{Question prompt and submission panel}
    \end{subfigure}
    \begin{subfigure}[t]{.9\textwidth}
        \includegraphics[width=\linewidth]{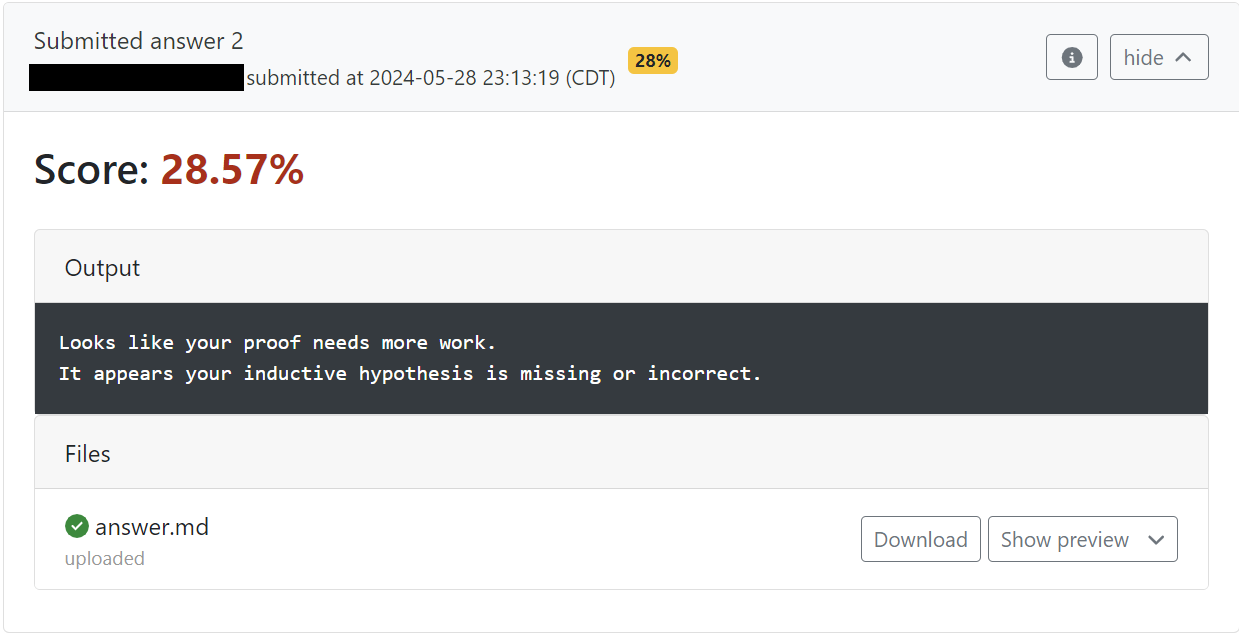}
        \caption{Graded results and feedback from the autograder}
    \end{subfigure}
    \caption{An example of an autograded problem in PrairieLearn. Students type in 
    their proof in the markdown editor box, and are able to see the markdown 
    preview in real time. Once they click ``Save \& Grade'', they receive 
    immediate feedback form the autograder, indicating if the submission is 
    correct or not. }
    \label{fig:pl-problem}
\end{figure}

Using the trained grading pipelines, we implemented an autograder for proof by 
induction problems using the online assessment system PrairieLearn~\cite{prairielearn}. Due to computational 
resource restrictions and the comparable performance of GPT-3 to Llemma 
on the 3 proofs used in our user study, we chose to use GPT-3 as the 
embedding model for our user study. For each student-submitted proof on a problem, 
the grading pipeline evaluates the proof using the rubrics R1-R7 introduced 
in~\ref{sec:data}, producing 7 correct/incorrect results. 
The final score for each submission is the sum of the 7 rubric points, ranging between 0 and 7, 
and then rescaled to a percentage between 0-100. 
\new{For each submission, we first show a general feedback based on the proof's percentage score, 
such as “you are almost there” or “your proof needs more work”, encouraging students 
to revise their submissions. Additionally, each rubric point is mapped to a specific 
feedback sentence that is shown to the student when a mistake is identified for that rubric 
point. For instance, an error in rubric item R4 (setting the bound of the inductive hypothesis) 
generates the feedback: “It appears your inductive hypothesis is missing or incorrect.” 
According to the taxonomy of feedback introduced in Shute~\cite{shute2008feedback}, the feedback 
provided 
by our grading pipeline is partially elaborated. We tell the students a rubric point which their 
proof is failing, but we do not explain to them how to fix it.
Figure~\ref{fig:pl-problem} shows an example page from PrairieLearn where students can submit 
their proofs and receive automatic feedback. 
We did not use generative 
models for feedback due to concerns about their accuracy at the time of the study design, and 
because we felt that rubric-point based feedback would be sufficiently elaborated for students 
to improve their proofs. As we will discuss more in Section~\ref{sec:feedback-helpfulness}, 
some students felt that the feedback was sufficiently helpful while others felt that it should 
have been more elaborated.
}

\subsection{Study Setup}

To investigate the accuracy of our grading models in a real class and better 
understand if they can help students write better proofs without the support of 
human graders, we conducted a study in Spring 2024 with students registered in the 
Discrete Mathematics course (CS173) at the University of Illinois 
Urbana-Champaign. 
We address the following research questions: 

\vspace{1em}
\begin{itemize}
    \item[\textit{RQ1}] Are students able to write better proofs by interacting with the autograder and the feedback it generates?
    \item[\textit{RQ2}] Are students satisfied with the autograder and the feedback it provides?
    \item[\textit{RQ3}] Does using the autograder make students more willing to 
    use similar AI autograders in the future? 
\end{itemize}
\vspace{1em}

The study was conducted earlier in the semester before students had been introduced to systematic instructions on proof by induction or had practiced any induction problems at the college level. 
As part of the study, students were asked to complete a learning activity and were offered extra credit points for its completion, equivalent to the completion of one homework (about 0.36\% of the course grade). 
The students had one week to complete the learning activity at a time of their choice.

The learning activity consisted of three parts. In the first part, students were 
expected to read a 6-page textbook section about induction, including information 
on the theoretical ground for induction, the steps of an induction proof, and an 
example problem with solution. The second part asked students to complete 3 
induction problems (P1-P3 as defined in~\ref{sec:data}). We only consider 3 
problems out of the 4 used in the model training process for two reasons: first, 
we wanted to limit the length of the learning activity to 1 hour, which would not 
be reasonable if we expected students to read the text and complete all 4 
problems; second, we removed P4 due to its low accuracy compared to P1-P3, as 
this could potentially hinder the students' learning experiences. 
The third part of the learning activity included a feedback survey, with the following questions: \vspace{2mm}
\begin{itemize}
    \item[\textbf{S01}] In CS173, I usually receive accurate grading from human graders for my proofs.
    \item[\textbf{S02}] Feedback from human graders has helped me improve my proofs.
    \item[\textbf{S03}] Overall, I am satisfied with my experience with human graders in CS173.
    \item[\textbf{S04}] I would prefer to wait for a week to get human grading feedback for the proofs I write for my homework than use the autograder for instant feedback.
    \item[\textbf{S05}] Even if a well-developed AI autograder has about the same accuracy as human graders, I still trust the grading results from human graders more.
    \item[\textbf{S06}] I am comfortable using a well-developed AI autograder in my course to give me feedback as I prepare for my quizzes.
    \item[\textbf{S07}] Given that I can still regrade my work with human graders, I am comfortable having a well-developed AI autograder in my course to grade my quizzes.
    \item[\textbf{S08}] I received accurate grading from the autograder.
    \item[\textbf{S09}] Feedback from the autograder helped me improve my proofs.
    \item[\textbf{S10}] Overall, I am satisfied with my experience of using the autograder.
\end{itemize} \vspace{2mm}

The first subsection of the survey includes questions S01-S03, which asks the 
students about their prior experience with human graders in the course. The second 
subsection includes questions S04-S07, focusing on students' perceptions of using 
AI for autograding. The last questions ask students to share their experiences 
with the AI autograder during the learning activity. 
The survey questions are presented on a five-point Likert scale using the following options and corresponding numeric values: ``Strongly disagree'' (-2), ``Disagree'' (1), ``Neutral'' (0), ``Agree'' (1), ``Strongly agree'' (2). Negatively worded questions (e.g., ``I would prefer to wait for a week to get human grading feedback for the proofs I write for my
homework than use the autograder for instant feedback'') are reverse coded in the 
statistical analysis. 

To get a better understanding of student perceptions of the graders, we 
included the following open-ended survey questions: 
\vspace{2mm}
\begin{itemize}
    \item[\textbf{S11}] In your opinion, how can the autograder be useful in your 
    learning experiences?
    \item[\textbf{S12}] Can you share a positive experience you had with the 
    autograder?
    \item[\textbf{S13}] Can you share a negative experience you had with the 
    autograder?
    \item[\textbf{S14}] Do you have other comments on the autograder?
\end{itemize} \vspace{2mm}

We do not use S11-S14 to directly answer any of the research questions, but we 
will use them in the discussion as we seek to understand the student's reasoning 
behind why they answered the way they did on the Likert items.

\subsection{Experimental Conditions}

The students were randomly assigned to 3 groups. We refer to the groups as 
\textit{Self-eval}, \textit{First}, and 
\textit{Random}. The \textit{Self-eval} group was the control group.
In the \textit{Self-eval} group, students did not have access to the AI autograders during the learning activity. 
Instead, upon proof submission, students in this group could see the 7 rubric 
points for the problem and were asked to self-assess and make improvements to 
their proof until they were satisfied with the solution. This is our baseline 
group, as it mimics a student completing a proof on their own.
 
Both the \textit{First} and \textit{Random} groups had access to the AI autograders, 
forming our treatment groups. Students in the \textit{First} group received 
feedback on the first rubric point (out of the seven) that the autograder identified. 
The order of these rubric points follows the logical sequence of R1–R7, 
corresponding to the typical flow of an induction proof. For example, an induction proof 
commonly begins with stating and proving the base cases, followed by stating the goal 
for the induction steps, and then completing the steps themselves.
The strategy of reporting the first incorrect rubric point is based on some students' preference to work incrementally, fixing the earlier parts of their work before moving on to the next part. 
When we designed the experiment, we expected this strategy to be the most helpful to students.

Students in the \textit{Random} group received feedback for one of rubric points, 
randomly selected from the rubric items the autograder identified as incorrect. 
\new{This approach was included in our study to explore whether different feedback strategies yield 
varying effects. The random strategy aimed to expose students to a broader range of 
feedback on the question, potentially helping them identify actionable steps to correct 
their submission. This approach could be especially beneficial for students who 
might otherwise remain stuck on the same part of the proof for an extended period of time.}

\section{User Study Results} \label{sec:user_results}

After the learning activity deadline, we manually screened all students' 
submissions and excluded students who showed no effort during the study. These 
excluded students either submitted blank proofs to all 3 problems or only typed 
something trivial such as ``hello'' or a question mark in their proofs. 
After applying this exclusion criteria, we had 68 students in the 
\textit{Self-eval} group, 59 students in the \textit{First} group, and 42 students 
in the \textit{Random} group. Due to constraints imposed by the online assessment platform PrairieLearn, it was
easier to pre-assign all eligible students to one of the experimental conditions 
before they elected to participate. Students agreed to participate upon opening 
the assessment in PrairieLearn. These two exclusion criteria combined resulted in 
the small variation in the population of sizes for each treatment. This small 
variation has no effect on the conclusions of the study because the assignment of 
students to experimental groups was still random.

\subsection{\textit{RQ1:} Are Students Able to Improve Their Proofs Using The Autograder?}

\begin{figure}
    \centering
        \includegraphics[width=.9\textwidth]{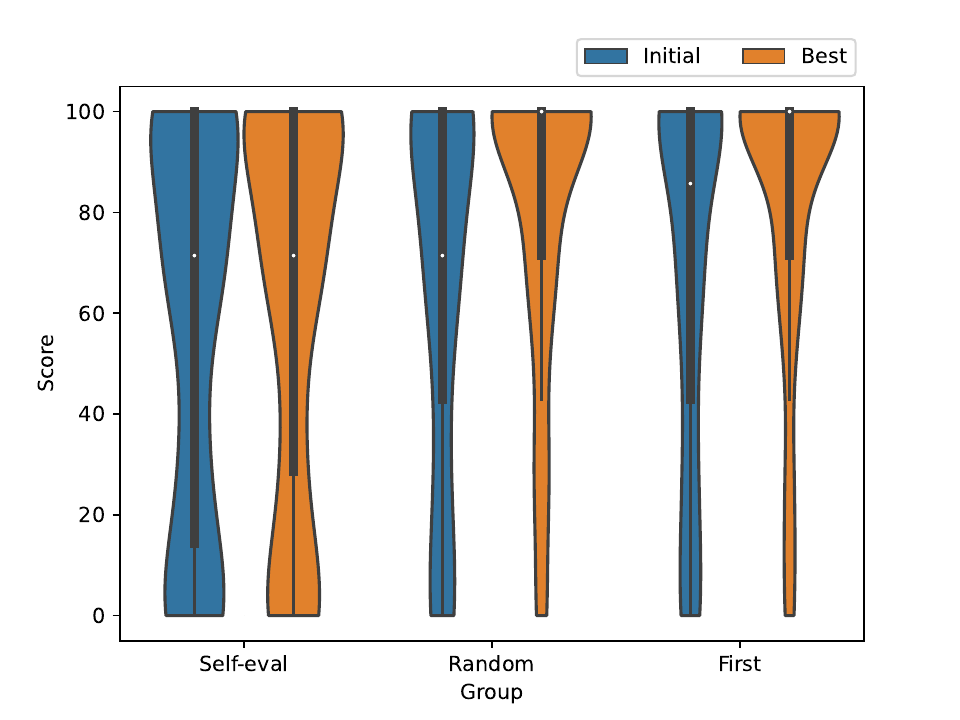}
        \caption{Distribution of initial and best scores in all problems, separated by the group. Distributions of initial and best submissions for the \textit{Self-eval} group are similar, while \textit{Random} and \textit{First} have different distributions. }
        \label{fig:scores}
\end{figure}

\begin{figure}
    \centering
        \includegraphics[width=.9\textwidth]{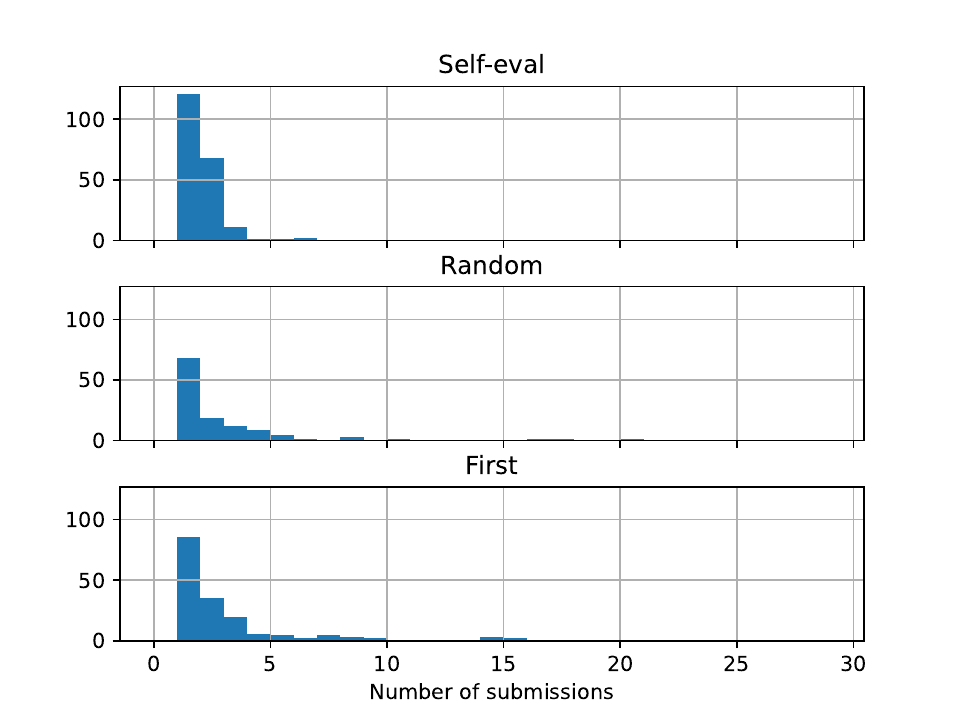}
        \caption{Histogram for the number of submissions per student for each of the three groups. Students in the \textit{Self-eval} group have slightly fewer submissions for all proof problems combined. }
        \label{fig:num-submissions}
\end{figure}

For each student, we collected the number of submissions for each problem and the score for each submission. 
We are interested in investigating the difference between students' initial submissions and best submissions for all three problems. 
Figure~\ref{fig:scores} shows the distribution of the initial scores and best scores for each student group, and Figure~\ref{fig:num-submissions} shows the number of submissions for all the proof problems combined. 

For each of the three proof problems (P1-P3), we performed a Kruskal-Wallis test across all groups to find the difference between students' earned scores on their initial submission. 
The results are $p=0.26, 0.14, 0.27$ respectively. 
These indicate that students are starting their proof-writing with similar quality across all groups, which makes sense as the students are randomly assigned to the three groups. 

On the other hand, a Kruskal-Wallis test on students' earned scores on their best 
submission shows statistically significant differences for all three problems, as 
shown in Table~\ref{tab:best_kruskal}. Post-hoc pairwise tests reveal that the 
best submission scores are significantly higher for students in \textit{First} and 
\textit{Random} than \textit{Self-eval}, while there is no significant difference 
between \textit{First} and \textit{Random}. 

\begin{table}[h]
    \centering
    \begin{tabular}{c c c c c c}
        \cmidrule(lr){2-4} 
    & \multicolumn{3}{c}{Mean(SD)} && \\ [-0.2em] \cmidrule(lr){2-4} 
    Problem & Self-eval & Random & First & $H$ & $p$ \\ [0.2em] \hline  
    P1 & 75.8(38.8) & 90.1(24.1) & 92.1(22.9) & 10.95 & 0.004 \\ [0.1em]
    P2 & 60.1(32.4) & 79.1(24.6) & 76.8(25.6) & 13.62 & 0.001 \\ [0.1em]
    P3 & 51.1(40.4) & 68.8(37.8) & 75.9(34.6) & 13.68 & 0.001 \\
    \end{tabular}
    \caption{Kruskal-Wallis, mean, and standard deviation for the best submission scores of each problem. Students in \textit{Random} and \textit{First} have a significantly higher score in their best submissions. }
    \label{tab:best_kruskal}
\end{table}

To better understand the effect of using the AI autograder to support the writing 
of proofs while controlling for initial student knowledge, we propose a regression 
model to determine the score gain for students in the three groups, using the 
students' initial and best submission scores. We fit an ordinary least squares 
(OLS) model of the form 
\begin{equation}
    \label{regression_model_performance}
        \mathrm{BEST}_{ij} = 
        \mu_j + \alpha \, \mathrm{I}_{ij} + 
        \beta_1 G1_i + \beta_2 G2_i
\end{equation}
where $\mathrm{BEST}_{ij}$ is the predicted best submission score for student $i$ on problem $j$, and $\mathrm{I}_{ij}$ is the initial score for student $i$ on problem $j$. Both scores range between 0 and 100. 
$G1_i\ \mathrm{and}\ G2_i$ are indicators of the students being in \textit{Random} or \textit{First}, respectively. For any student in \textit{Self-eval}, both values are 0; for students in \textit{Random} $G1_i$ is 1 and $G2_i$ is 0, and for students in \textit{First} $G1_i$ is 0 and $G2_i$ is 1. 
We want to estimate the parameters $\mu_j, \alpha, \beta_1, \mathrm{and}\  \beta_2$, which can be interpreted as follows: 

\begin{itemize}
    \item $\mu_j$: Control for the difficulty of problem $j$
    \item $\alpha$: Control for the initial score of the proof
    \item $\beta_1$: The effect of feedback for students in the \textit{Random} group
    \item $\beta_2$: The effect of feedback for students in the \textit{First} group
\end{itemize}
\hspace{.1em}

Table~\ref{tab:regression_1} summarizes the results from the regression analysis using Equation~\ref{regression_model_performance}. 
The baseline students are students in \textit{Self-eval}, who do not receive the results or feedback from the autograder. 
Compared with \textit{Self-eval} students, students in the \textit{First} and \textit{Random} groups are able to gain respectively 11.6 and 11.3 more points from the proofs they write after controlling for their initial submission performances. 
This shows that students can improve and write more accurate proofs using the 
feedback from the autograder compared with students who are just self-evaluating 
using the rubric. In future work, we will seek to also measure knowledge retention 
by having students complete additional proofs without the autograder in a delayed 
posttest.

\begin{table}[h]
    \centering
    \begin{tabular}{c | c | c c c c}
        Coefficient & Description & Value & $std err$ & $t$ & $p$ \\ \hline
        $\alpha$ & Control for initial score & 0.693 & 0.03 & 27.9 & $<0.001$ \\ 
        $\beta_1$ & Effect of the \textit{Random} group & 11.3 & 2.32 & 4.86 & $<0.001$ \\ 
        $\beta_2$ & Effect of the \textit{First} group & 11.6 & 2.10 & 5.51 & $<0.001$ \\ 
        $\mu_1$ & Control for difficulty of P1 & 26.5 & 2.56 & 10.36 & $<0.001$ \\ 
        $\mu_2$ & Control for difficulty of P2 & 20.2 & 2.37 & 8.52 & $<0.001$ \\
        $\mu_3$ & Control for difficulty of P3 & 20.0 & 2.25 & 8.92 & $<0.001$ \\
        
    \end{tabular}
    \caption{Coefficients from Equation~\ref{regression_model_performance}. The $R^2$ metric of this regression is 0.665, indicating a strong relationship between variables. Students in the experimental groups using the AI autograder are able to achieve 11 more points in the proofs than students using the rubrics for self-evaluation. }
    \label{tab:regression_1}
\end{table}



\subsection{Survey Results} \label{sec:survey_results}

\begin{figure}[h]
    \centering
        \includegraphics[width=.9\linewidth]{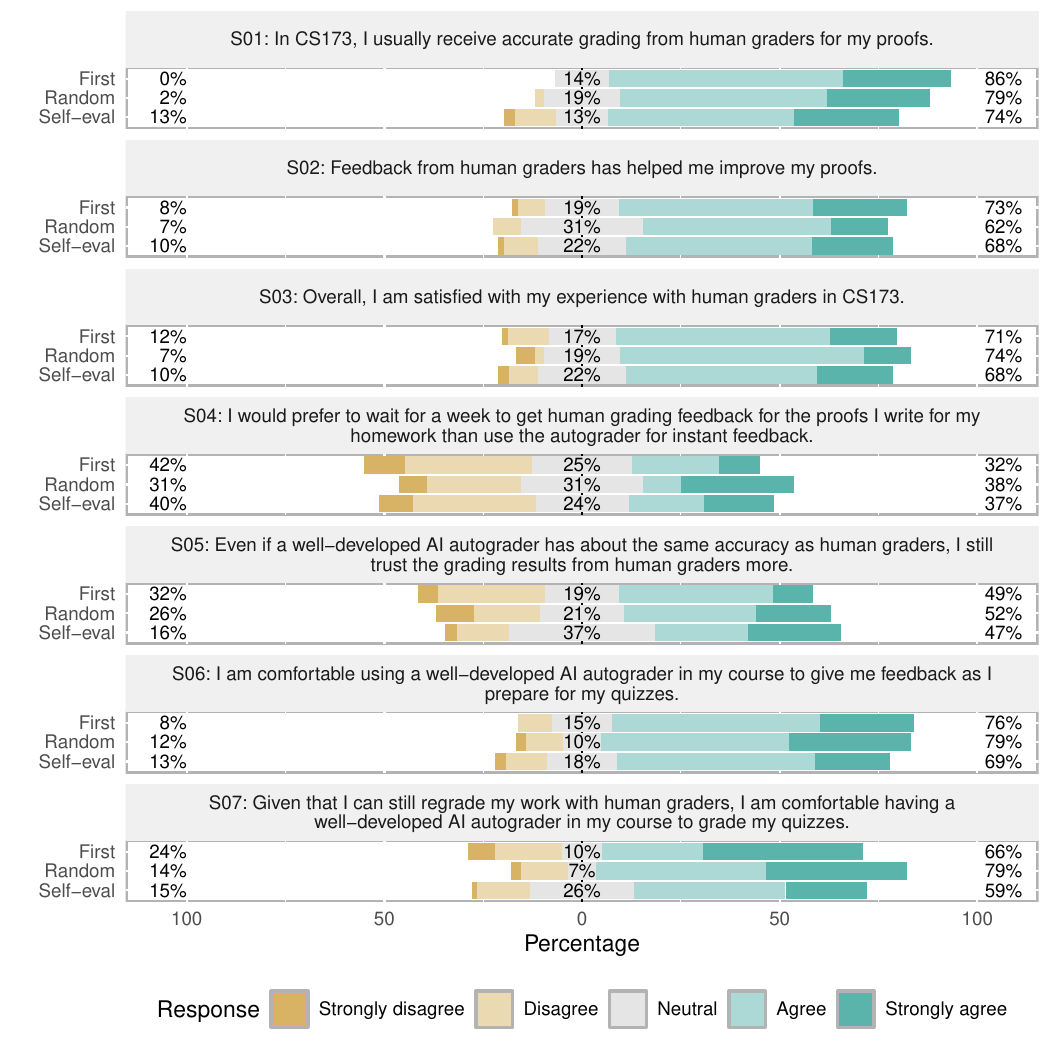}
        \caption{Responses from survey questions given to all 3 groups. The percentages shown on each distribution represent the percentage of negative responses (disagree \& strongly disagree), the percentage of netural responses, and the percentage of positive responses (agree \& strongly agree). }
        \label{fig:survey_1}
\end{figure}

\begin{figure}[h]
    \centering
        \includegraphics[width=.9\linewidth]{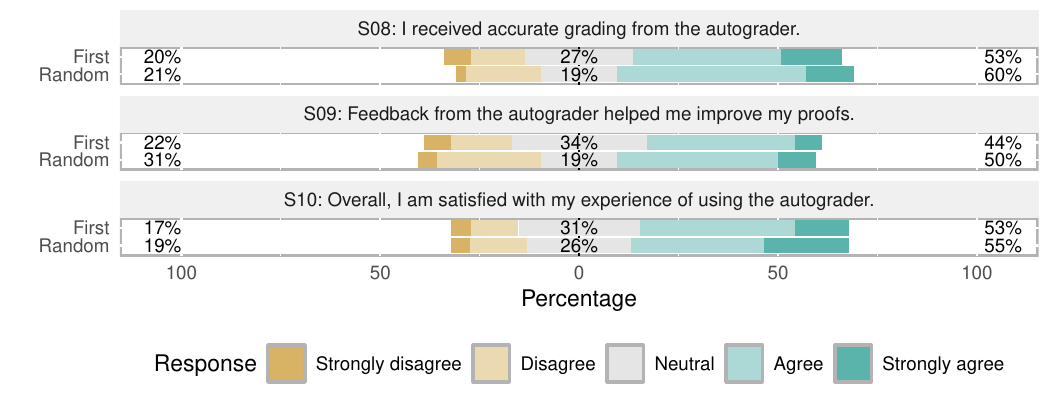}
        \caption{Responses from survey questions given to only \textit{First} and \textit{Random}. The percentages shown on each distribution represent the percentage of negative responses (disagree \& strongly disagree), the percentage of neutral responses, and the percentage of positive responses (agree \& strongly agree).}
        \label{fig:survey_2}
\end{figure}

The survey results are shown in Figure~\ref{fig:survey_1} and 
Figure~\ref{fig:survey_2}. The three subsections of the surveys address different 
aspect of student perceptions: S01-S03 
address student perceptions of human graders,  
S04-S07 address students willingness to have AI graders integrated into various 
parts of their course, and S08-S10 address student perceptions of the AI graders. 
We calculated a Cronbach's $\alpha$ for each of the three subsections. 
The calculated Cronbach's $\alpha$ are 0.81, 0.72, 0.82 respectively, indicating good internal reliability. 
The \textit{Self-eval} group were not asked questions S08-S14 as these students 
did not interact with the autograder during the learning activity. 

\subsubsection{\textit{RQ2:} Are students satisfied with the autograder and 
feedback?}
We use responses from survey questions S01-S03 and S08-S10, as shown in Table 
~\ref{tab:survey_RQ1}, to address RQ2. S01-S03 capture students' perceptions of 
human graders, while S08-S10 assess their perceptions of the AI autograder. 
Specifically, S01 and S08 measure the perceived accuracy of the graders, S02 and 
S09 evaluate their helpfulness, and S03 and S10 measure overall satisfaction with 
the graders.
On average, human graders are perceived to be more accurate and helpful than the 
AI autograder, with   statistically significant differences ($p<0.05$). However, 
although the mean satisfaction rating is higher for human graders, the differences 
are not statistically significant for either the First or Random groups. We 
hypothesize that this is due to the known benefits of the AI autograder, such as 
timely feedback and the ability to submit an unlimited number of times.

\begin{table}[h]
    \centering
\begin{adjustbox}{width=\columnwidth}

    \begin{tabular}{c | c c c c | c c c c | c c c c}
        & \multicolumn{4}{c |}{Accuracy} & \multicolumn{4}{c |}{Helpfulness} & \multicolumn{4}{c}{Satisfaction} \\ 
        Group 
        & S01 & S08 & $t$ & $p$ 
        & S02 & S09 & $t$ & $p$ 
        & S03 & S10 & $t$ & $p$ \\ \hline
        Self-eval 
        & 0.84 & --- & --- & ---
        & 0.76 & --- & --- & ---
        & 0.74 & --- & --- & --- \\
        First 
        & 1.14 & 0.41 & 4.37 & $<0.001$
        & 0.86 & 0.22 & 3.61 & $<0.001$
        & 0.75 & 0.44 & 1.69 & 0.094 \\
        Random 
        & 1.02 & 0.48 & 2.81 & 0.006
        & 0.69 & 0.24 & 2.14 & 0.035
        & 0.74 & 0.52 & 0.97 & 0.336 \\
        
    \end{tabular}
\end{adjustbox}
    \caption{S01-S03 and S08-S10 statistics for three groups, and the pairwise t-test for independence. \textit{Self-eval} students did not have access to S08-S10 as these students did not interact with the autograder during the learning activity. All scores range between -2 and 2. }
    \label{tab:survey_RQ1}

\end{table}

\subsubsection{\textit{RQ3:} Does using the autograder make students more willing 
to use AI autograders?}
We aim to understand students' perceptions of and trust in using an autograder 
for their coursework across various settings. Specifically, we want to determine 
whether interactions with the autograder can foster more positive perceptions.
These are reflected through S04-S07 (see Table~\ref{tab:survey_RQ2}). 
S04 examines students' preferences between receiving instant feedback from AI autograders and obtaining more reliable feedback from human graders. S05 assesses the level of trust students have in AI autograders in general. S06 and S07 explore potential applications of AI autograders. 

\begin{table}[h]
    \centering
    \begin{tabular}{c | c c c c c}
        Survey & Self-eval & Random & First & $F$ & $p$ \\ \hline
        S04 & -0.06 & -0.29 & 0.10 & 1.20 & 0.31 \\ 
        S05 & -0.51 & -0.36 & -0.22 & 1.06 & 0.35 \\
        S06 & 0.72 & 0.95 & 0.92 & 1.01 & 0.37 \\
        S07 & 0.63 & 0.98 & 0.76 & 1.17 & 0.31 \\
        
    \end{tabular}
    \caption{S04-S07 mean numeric scores about perceptions using AI autograders, with one-way ANOVA test statistics. All scores range between -2 and 2. }
    \label{tab:survey_RQ2}
\end{table}

On average, students in \textit{First} and \textit{Random} are slightly more comfortable using a well-developed autograding system in the future. 
However, one-way ANOVA tests among the groups show the differences are not statistically significant. 

\section{Discussion}

\subsection{Model Accuracy}
In terms of student-perceived accuracy, the results mostly agree with the accuracy measures in~\ref{sec:human_results}. Although GPT-3 models are observed to perform relatively well, they are still less accurate than experienced human graders. 
This difference in accuracy can be amplified even more in practice. 
Repetition of similar proofs might skew the accuracy measures because similar proofs tend to be graded the same way due to the stable nature of the large language models. 
Students can potentially encounter multiple incorrect gradings in a row by submitting similar proofs, which increases their sense of mistrust on the graders. 

Currently, the large language models are used as black boxes for embedding. There are cases where the models do not precisely interpret the input sentences, but it is hard to predict why, when, or how the misinterpretation happens. This can lead to the autograder giving wrong grades in mysterious ways. 
This has happened to some of the users. Some open-ended feedback from the users are listed below: 
\begin{itemize}
    \item ``Additionally, when I added `Suppose that' in front of my Inductive Hypothesis, it deemed it incorrect.''
    \item ``It was not always accurate especiall [sic] when I didnt us LateX perfectly.''
    \item ``My grade also improved by changing some, but not all, of my `\verb|\times|' to `\verb|\cdot|' for no apparent reason, either. When I changed all of them, my grade remained unchanged.''
    \item ``Spacing my sentences out into separate paragraphs improved my grade by 15\%, also for no apparent reason.''
\end{itemize}
One possible improvement for this situation is to break down a proof problem into 
sub-problems, where students submit portions of the proof into different input 
boxes. 
This can help decrease the complexity of the proofs being submitted and provide 
more accurate embeddings. 
This also makes the task slightly different due to the added scaffolding, which 
may be appropriate for formative settings, but less so for exams.

\subsection{Feedback Helpfulness}
\label{sec:feedback-helpfulness}
\new{The elaborated nature of the feedback from our grader enabled students using the autograder to make good progress on the proofs, 
and being able to see the feedback almost instantly also provided students a better way of practicing writing the proofs.}
\begin{itemize}
    \item ``The autograder tells me where exactly I am lacking information, which helps me more because I can pinpoint where I need to work on.''
    \item ``The autograder allows for quick and efficient feedback of on proofs versus waiting to hear back days after on issues with ones proof. This quick turn around time allows for a easy/fast way to get better at proof writing.''
    \item ``I got absolutely stuck with the last proof! However, when I used the autograder, it gave me around a 75\%. This gave me confidence that I was on the right track, and I was able to not get discouraged. I was slowly able to incorporate more parts of the solution into my proof''
\end{itemize}

\new{On the other hand, more elaborated feedback or instructions such as hints to fix the proofs would help student make even more improvements. }
\begin{itemize}
    \item ``Some of the feedback was basic. didn't say anything specific. ''
    \item ``I also think that the autograder was not specific enough to identify what exactly was wrong with my proofs. I was stuck on the 2nd problem for a while, and the feedback that I was getting was not exactly enough to pinpoint what exactly was wrong.''
\end{itemize}
\new{Providing such automated and elaborated feedback should be possible with the help of generative models. 
We did not use generative models for this study due to concerns about low grading accuracy resulting in low feedback accuracy. 
At the time of the study, generative AI models stuggled to give feedback on long technical passages such as the induction proofs. 
However, they have improved over time so we can try them in future work.} 
By leveraging few-shot training techniques such as Retrieval-Augmented 
Generation~\cite{lewis2021retrieval-augmented} and including the more accurate grading outputs from our models, we can fine-tune chatbots such as 
ChatGPT to provide suggestions or hints with reduced likelihood of hallucination.  
These fine-tuned chatbots can also better align with the specific learning objectives of individual courses. This 
potential enhancement to our system can be explored in future work, allowing us 
to get a better understanding into the quality of AI-generated feedback. 

\subsection{Perceptions on AI grading}

Proof grading is a subjective task, which can lead to low levels of trust from students. 
When designing the learning activities, we initially hoped that the interactions with our AI autograder would increase students' trust and willingness to use such tools in the future.
However, results in~\ref{sec:survey_results} do not confirm our initial hypothesis. 
Students in \textit{First} and \textit{Random} are slightly more willing to use an AI autograder in various settings, but the difference is not statistically significant. 
Simply using the autograder does not sufficiently increase students' willingness 
to use AI autograders in general. 
One potential strategy to build trust in AI autograders is to provide more detailed feedback 
and explanations during the grading process. Research by Ha and Kim~\cite{ha2023improving} indicates 
that offering additional information about AI models can be effective in reducing cognitive biases 
and fostering trust. Incorporating such measures into future iterations of the autograder could 
improve its acceptance and perceived reliability among students.

\section{Conclusions and Future Work}
In this study, we developed a set of machine learning models to autograde freeform mathematical proofs. These models enable fast training \new{without the need for extensive fine-tuning}, and they achieve accuracy comparable to that of most human graders recruited for this study. 
These models can be used to develop NLP graders which could be deployed on large-scale educational platforms in the future, enabling more students to receive the necessary support when learning to write mathematical proofs. 

After creating and deploying autograded proof by induction problems using the 
grading models, we conducted a user study to evaluate the autograder's impact. 
Our results show that students who used the autograder for feedback to make revisions to their submissions scored over 10\% higher on the proof 
problems than students who did not use the autograder. This is an encouraging sign for 
creating more autograded problems in the future. Despite being helped by the 
autograder, many students still preferred feedback from human graders, even if it was delayed.

In the future, fostering greater student trust in AI autograders will be an important area of focus.
Efforts can be put on improving the grading models as well. Since our current models were trained exclusively on mathematical induction problems, their ability to grade other types of proof such as proof of bijection or geometry is still unknown. We plan to collect data on various types of mathematical proof problems, and check if our grading models can perform equally well on these problems. 

\new{Additionally, he cost of training and running these models can also be reduced. 
In our current design, GPT-3-based models seem to be the best choice as they can be scalable at a low cost and achieve a relatively good performance. 
However, there is still a lack of open-source language models that are computationally efficient, 
cost-effective, and capable of achieving comparable performance. Developing such models could allow for greater customization and broader application of AI autograders, 
making this an important focus for future work with more advanced language models.}

\subsubsection*{Acknowledgments} 
This work used the Delta system at the National Center for Supercomputing 
Applications through allocation CIS230355 from the Advanced Cyberinfrastructure 
Coordination Ecosystem: Services \& Support (ACCESS) program, which is supported 
by National Science Foundation grants \#2138259, \#2138286, \#2138307, \#2137603, 
and \#2138296. 

The work is also made possible by members of previous research projects collecting and 
labeling the proof data, and graders from multiple institutes providing grading 
data for comparison.  


\bibliography{references}

\end{document}